\documentclass{article}

\usepackage{arxiv}

\usepackage[utf8]{inputenc} % allow utf-8 input
\usepackage[T1]{fontenc}    % use 8-bit T1 fonts
\usepackage{hyperref}       % hyperlinks
\usepackage{url}            % simple URL typesetting
\usepackage{booktabs}       % professional-quality tables
\usepackage{amsfonts}       % blackboard math symbols
\usepackage{nicefrac}       % compact symbols for 1/2, etc.
\usepackage{microtype}      % microtypography
\usepackage{lipsum}         % Can be removed after putting your text content
\usepackage{graphicx}
\usepackage[numbers]{natbib}
\usepackage{doi}

\usepackage{amsmath}
\usepackage{amssymb}
\usepackage{amsthm}
\usepackage{mathtools}
\usepackage{bm}
\usepackage{cleveref}       % smart cross-referencing

\title{Graph Neural Networks for Automatic Addition of Optimizing Components in Printed Circuit Board Schematics}

\date{June 12, 2025}

\usepackage{authblk}

\author[1]{%
	Pascal Plettenberg\thanks{\texttt{plettenberg@uni-kassel.de. \href{https://orcid.org/0000-0003-4934-7382}{https://orcid.org/0000-0003-4934-7382}}}}%
\author[2]{%
	André Alcalde}%
\author[1]{%
	Bernhard Sick}%
\author[1]{%
	Josephine M. Thomas}%
\affil[1]{Intelligent Embedded Systems, University of Kassel, 34121 Kassel, Germany}
\affil[2]{CELUS GmbH, 80339 München, Germany}

\renewcommand{\shorttitle}{GNNs for Automatic Addition of Optimizing Components in PCB Schematics}

\hypersetup{
pdftitle={Graph Neural Networks for Automatic Addition of Optimizing Components in Printed Circuit Board Schematics},
pdfsubject={},
pdfauthor={Pascal~Plettenberg, André~Alcalde, Bernhard~Sick, Josephine~M.~Thomas},
pdfkeywords={Graph Neural Networks, Printed Circuit Boards, Electronic Design Automation},
}

\begin{document}
\maketitle

\begin{abstract}
The design and optimization of Printed Circuit Board (PCB) schematics is crucial for the development of high-quality electronic devices. Thereby, an important task is to optimize drafts by adding components that improve the robustness and reliability of the circuit, e.g., pull-up resistors or decoupling capacitors. Since there is a shortage of skilled engineers and manual optimizations are very time-consuming, these best practices are often neglected. However, this typically leads to higher costs for troubleshooting in later development stages as well as shortened product life cycles, resulting in an increased amount of electronic waste that is difficult to recycle. Here, we present an approach for automating the addition of new components into PCB schematics by representing them as bipartite graphs and utilizing a node pair prediction model based on Graph Neural Networks (GNNs). We apply our approach to three highly relevant PCB design optimization tasks and compare the performance of several popular GNN architectures on real-world datasets labeled by human experts. We show that GNNs can solve these problems with high accuracy and demonstrate that our approach offers the potential to automate PCB design optimizations in a time- and cost-efficient manner. 
\end{abstract}

\keywords{Graph Neural Networks  \and Printed Circuit Boards \and Electronic Design Automation.}

\section{Introduction}
The development of high-quality electronic devices relies heavily on the design of Printed Circuit Board (PCB) schematics, which define all required components and their connections to each other. Usually, new PCB schematics undergo several design iterations, and first functional drafts are optimized based on experience according to some "best practices". Thereby, one important task for engineers is to add components like pull-up resistors or decoupling capacitors to the circuit, which reduce the failure risk and increase the robustness against external disturbances.\\
In most cases, these types of PCB design optimizations are implemented manually, which is time-consuming and error-prone. Since the additional components are not needed for immediate functionality, they are often neglected. High time pressure on engineers in the development process, and the shortage of skilled engineers in the market contribute to that as well. Consequently, non-optimal PCB schematics result in more frequent prototype iterations and higher troubleshooting costs in later development stages. Furthermore, they reduce the overall lifetime of the final product, resulting in an increased amount of electronic waste that is difficult to recycle.\\
Electronic Design Automation (EDA) tools are essential for streamlining the process of designing, testing, and verifying electronic designs. These tools have become even more sophisticated since Machine Learning (ML) methods have been integrated into EDA due to the increasing amount of available data \citep{huang2021machine}. While ML-based EDA tools have been mainly developed for integrated circuits (ICs), the automation of PCB schematics, which include both analog and digital components, is more difficult due to larger design spaces, and existing tools are often limited to rule-based design verifications.\\
Electronic circuits can be naturally represented as graphs and Graph Neural Networks (GNNs) have emerged as a powerful tool that extends the scope of Deep Learning from Euclidean to graph-structured data. Therefore, GNNs offer great potential for learning meaningful representations of PCB schematics that can be used to automate optimization tasks. However, one big problem is the translation of the huge variety of available information on the PCB components (e.g., types and names) into standardized, numerical node features, especially when dealing with real-world datasets. Furthermore, most GNN models from the literature are either used for prediction tasks on the level of single nodes \citep{hamilton2017inductive} or edges \citep{zhang2018link}, whereas adding a new component with exactly two terminals requires the classification of node pairs.\\\\
\textbf{Present work.} In this work, we propose a GNN-based approach for automating the placement of optimizing components in PCB schematics. Thereby, we represent PCB schematics as bipartite graphs (see Sec.~\ref{sec:graph_representation}) and predict the positions of new components with a node-pair-level classification model (see Sec.~\ref{sec:model}). In Sec.~\ref{sec:experiments}, we evaluate our approach on large real-world datasets labeled by human experts, thereby focusing on three specific PCB design optimization tasks, each involving the addition of a different optimizing component: (i) \textbf{Pull-up} and \textbf{pull-down resistors} for ensuring a defined voltage level on floating nets. (ii) \textbf{RC filters} on reset pins for preventing unintentional resets from voltage glitches. (iii) \textbf{Decoupling capacitors} for reducing high-frequency noise in supply and ground nets. We train our model to perform these tasks using different GNN architectures from the literature and compare their performances.\\\\
\textbf{Main Contributions}
\begin{itemize}
    \item[\textbullet] We propose a bipartite graph representation for PCB schematics involving a method for constructing node and edge attributes from non-standardized component names in real-world data using a pre-trained language model.
    \item[\textbullet] We propose a GNN-based node-pair-level prediction model for the placement of additional optimizing components in PCB schematics.
    \item[\textbullet] We perform extensive experiments on real-world datasets, demonstrating that GNNs can predict the positions of new components with high accuracy.
\end{itemize}
The code for our model and example graph samples from our dataset are available at \url{https://github.com/pasplett/pcb-node-pair-gnn}.

\section{Related Work}
\textbf{Machine Learning for Electronic Design Automation.} In the last years, many studies have investigated the usage of ML methods for EDA \citep{huang2021machine}. These studies mainly focused on digital circuit design, including several stages of the design flow such as logic synthesis \citep{yu2018developing,hosny2020drills}, placement \citep{mirhoseini2020chip}, and routing \citep{xie2018routenet,barboza2019machine}. In analog design, however, ML-based automation is more difficult due to larger design spaces and varying specifications \citep{mina2022review}. Some notable approaches are reinforcement learning for circuit topology optimization \citep{settaluri2020autockt} as well as ML-assisted analog circuit sizing \citep{budak2021efficient}. However, these approaches do not exploit the graph structure of electronic circuits.\\\\
\textbf{Graph Learning for Electronic Design Automation.} Since electronic circuits can be naturally represented as graphs, GNNs have recently become more and more popular in the field of EDA \citep{sanchez2023comprehensive}. Again, while the majority of studies focus on the digital EDA flow \citep{cheng2021joint,mirhoseini2021graph,shrestha2024eda}, there are also some GNN-based approaches for analog EDA, such as ParaGraph \citep{ren2020paragraph} for the prediction of net parasitic capacitances, GANA \citep{kunal2020gana} for automated netlist annotation and Circuit Designer \citep{wang2020gcn} for transistor sizing. Most recently, CktGNN \citep{dong2023cktgnn} was introduced as a nested GNN framework with a pre-designed subgraph basis and has been successfully applied for analog circuit topology design and device sizing. While all of these approaches utilize the graph structure of electronic circuits, they are tailored to different use cases and are not applicable to the specific PCB optimizations that we focus on in our work.\\\\
\textbf{Graph Neural Networks for Circuit Design Completion.} Closest to our work is the study by Said et al. \citep{said2023circuit}, which explores the usage of graph neural networks for the design completion of partially designed analog circuits. Thereby, the missing component is first identified using a graph classification, and the placement of the new component within the circuit is then treated as a link prediction problem. However, predicting the connectivity of a new component using link prediction frameworks like GAE \citep{kipf2016variational} or SEAL \citep{zhang2018link} is difficult because it involves the prediction of links to isolated nodes and the individual links have to be predicted independently. Here, we focus on adding new components for specific PCB schematic optimization tasks with practical relevance. Thereby, the number of connections for a new component is known in advance, so we can treat the problem as a node-pair-level prediction task rather than a combination of graph classification and link prediction. 

\section{Graph Representation of PCB Schematics} \label{sec:graph_representation}
\begin{figure}[t]
\centering
\includegraphics[width=\textwidth]{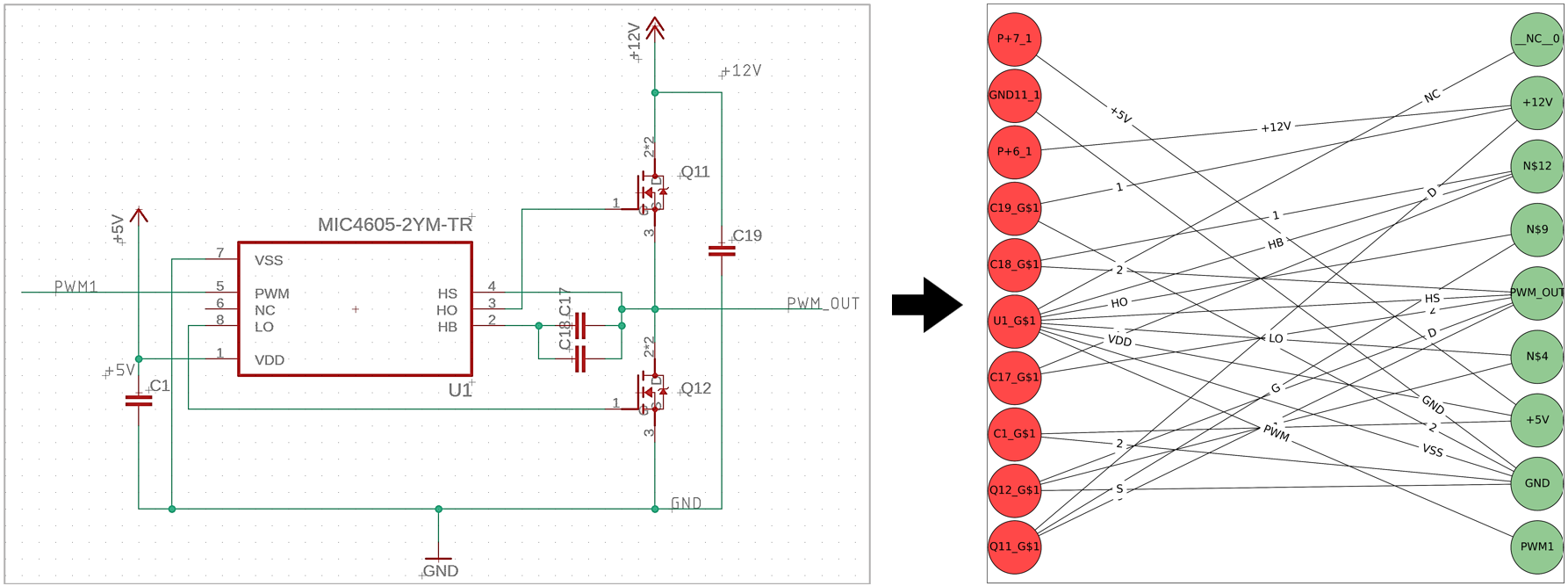}
\caption{An exemplary PCB circuit diagram and the corresponding bipartite graph representation. Symbols (e.g., resistors or capacitors) are represented by red nodes, nets (e.g., supply or ground nets) are represented by green nodes, and pins (e.g., component terminals) are represented by edges.} \label{fig:sch2graph}
\end{figure}

\begin{figure}[t]
\centering
\includegraphics[width=\textwidth]{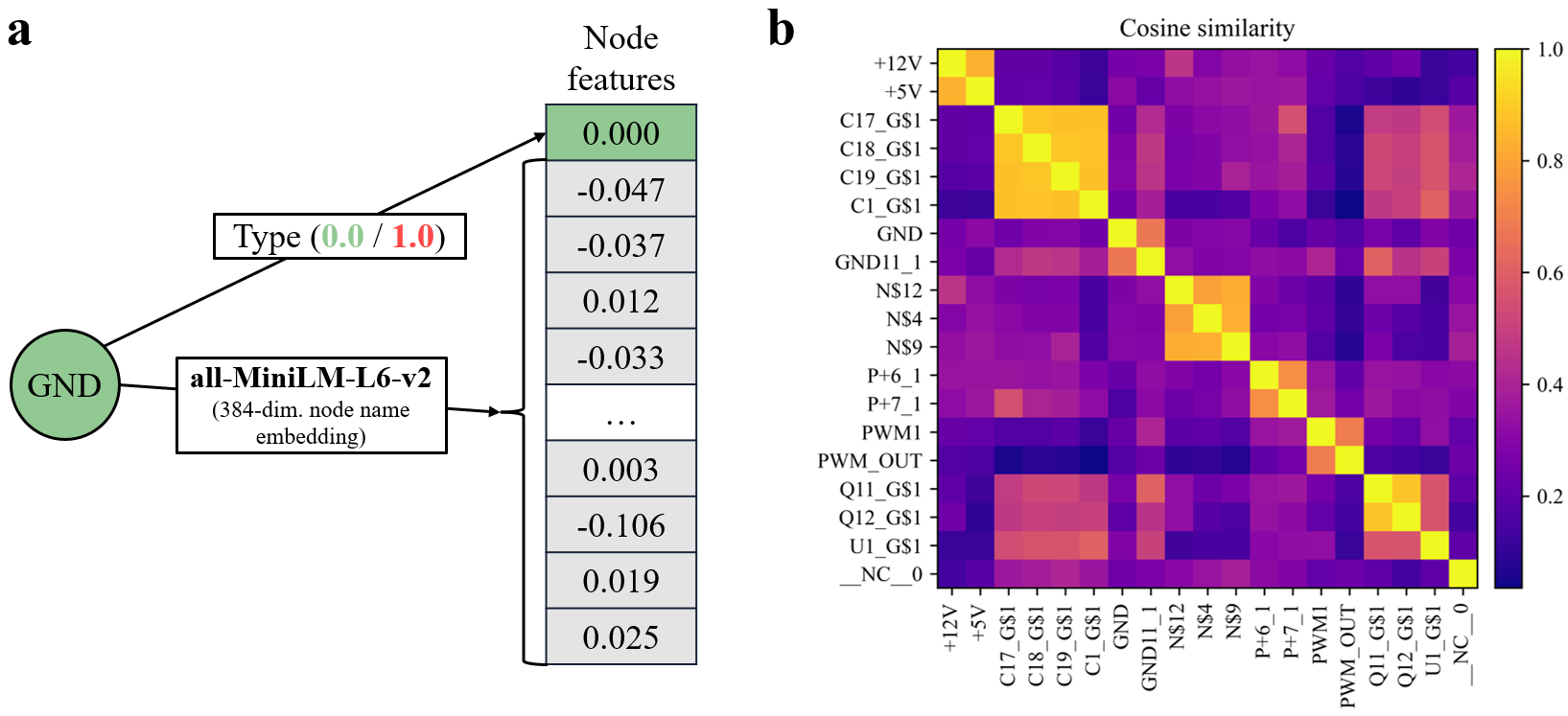}
\caption{\textbf{a} Composition of the input node features from node type and node name embedding. \textbf{b} Cosine similarity between sentence transformer embeddings of all symbol and net node names in an example PCB circuit diagram. Thereby, embeddings of similar node names exhibit higher cosine similarities.} \label{fig:node_features}
\end{figure}

We represent PCB schematics as bipartite graphs with two distinct sets of nodes: Nets and symbols. Thereby, net nodes (e.g., ground or supply nets) are always connected to symbol nodes (e.g., components like resistors, capacitors, or ICs) and vice versa, but no connections are allowed between nodes within the same set. The edges of the graph represent the pins, i.e., the terminals or connection points of the symbol nodes. Since a symbol may be connected to the same net node via multiple pins, the resulting graphs can be multi-relational, i.e., allow for multiple parallel edges. An example of the proposed graph representation of PCB schematics is given in Fig.~\ref{fig:sch2graph}.
\\\\
\textbf{Node Attributes.} The node input features contain two types of information: The type and the name of the node. The node type is a binary variable, which takes the value 0 for a net node and the value 1 for a symbol node. The node names are stored as strings within the raw schematics file, e.g. "GND" or "C1\_G\$1". We use a pre-trained language model to convert these strings into unique numerical embeddings. For this purpose, we utilize the model all-MiniLM-L6-v2 from the SentenceTransformers Python package \citep{reimers2019sentence}, which transforms any input string into a numerical vector containing 384 values. The model is a distilled version of a large Transformer model, which was trained on a diverse language dataset containing over 1 billion training pairs \citep{wang2020minilm}. The resulting node name embedding is concatenated to the single node type value resulting in the final 385-dimensional node input feature vector (see Fig.~\hyperref[fig:node_features]{\ref*{fig:node_features}a}).
\\\\
\textbf{Edge Attributes.} Similar to the node names, we also use the pre-trained sentence transformer model to convert the pin names into 384-dimensional edge attributes. Furthermore, since standard GNN architectures cannot properly handle multi-relational graphs, we combine parallel edges into a single edge by summing up the individual pin name embeddings and concatenating an additional edge feature representing the number of parallel pins represented by the final edge.\\\\
\textbf{Applicability to Real-World Data.} Our approach for representing PCB schematics as graphs is very flexible and requires a minimum amount of preprocessing, because the type of a PCB component, e.g., a resistor, capacitor, or IC, is implicitly encoded in the node features via the name embeddings. Therefore, identifying component types manually is not necessary, making the approach especially suitable for real-world datasets with non-standardized component names.  Fig.~\hyperref[fig:node_features]{\ref*{fig:node_features}b} shows the cosine similarity between the name embeddings of all nodes in the example PCB circuit diagram from Fig.~\ref{fig:sch2graph}. Note that small name variations (e.g., "C1", C17", "C18") result in similar sentence transformer embeddings, whereas nodes with very different names (e.g., "+5V" and "C17\_G\$1") exhibit a much lower cosine similarity.

\section{Node Pair Prediction for PCB Component Addition} \label{sec:model}
Our aim is to optimize the robustness and reliability of PCB designs by adding new relevant components and predicting their position within the circuit. In most cases, these new components are either resistors or capacitors, which is why we assume that the new circuit component is a two-terminal component, i.e., it is connected to exactly two net nodes. The task is to predict the pair of net nodes between which the new component has to be inserted.\\\\
\textbf{Node Pre-Filtering.} A straightforward approach would be to calculate node representations using GNNs and then predict a probability score for each possible pair of net nodes. However, in large circuits with hundreds of nodes, this approach may become computationally expensive because the number of possible node pairs increases quadratically. Furthermore, the approach could lead to training instabilities since the dataset can be extremely imbalanced: Only a few node pairs have a positive training label.\\ 
Therefore, we develop a strategy for reducing the number of node pairs that need to be checked during the prediction. First, we can restrict the search to net nodes because the inserted component is a symbol that cannot be directly connected to other symbol nodes. Second, we can sort out net nodes that are very unlikely to be connection points for the new circuit component. For example, the resistor of an RC filter is never connected to a ground net. This pre-filtering can be done with an MLP that predicts for each net node, whether it serves as a connection point for a new component. The resulting probability scores assigned by the pre-filter can be used to identify a set of unlikely candidates and exclude them from the actual pairwise node prediction.
\begin{figure}[!ht]
\centering
\includegraphics[width=0.9\textwidth]{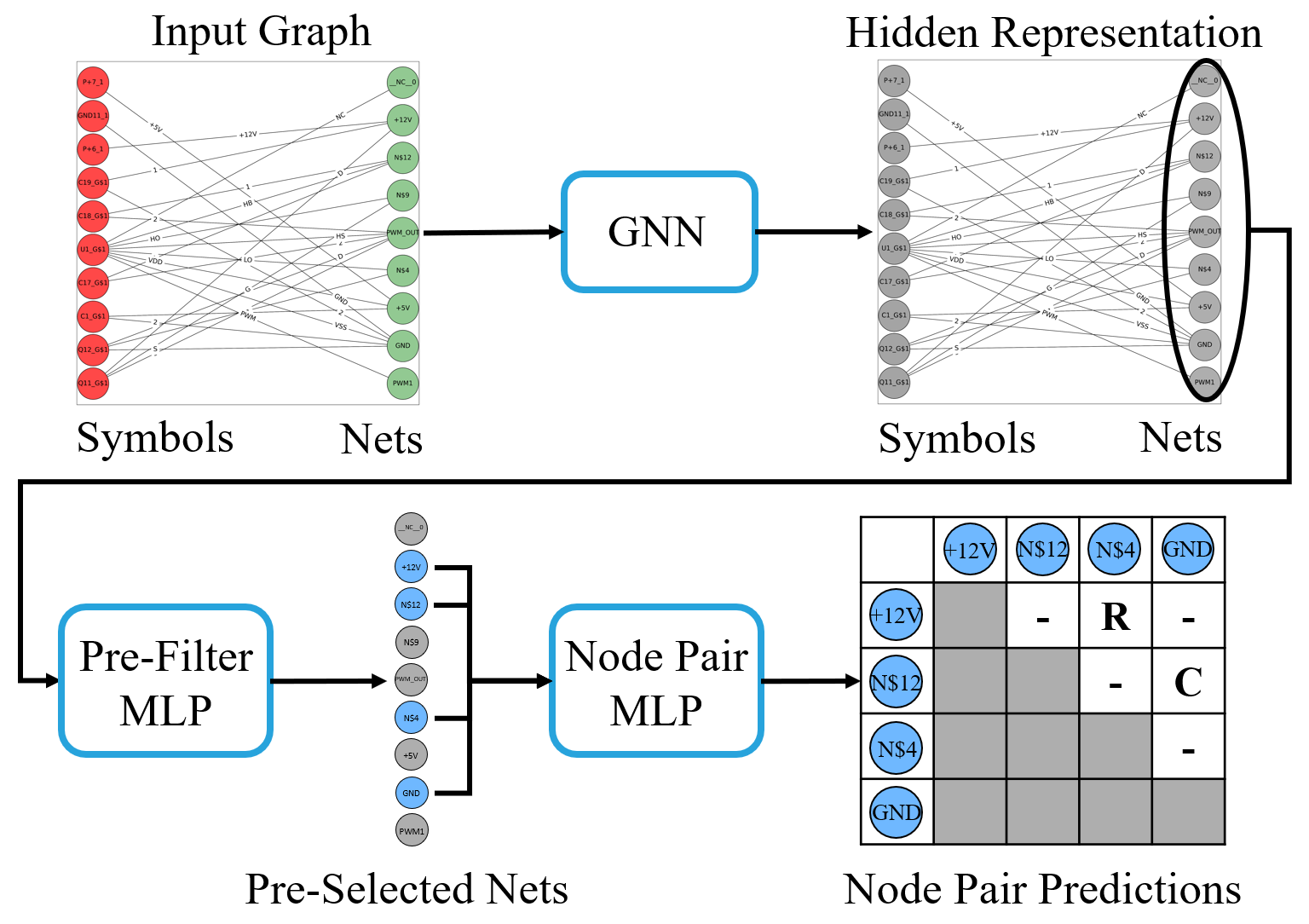}
\caption{Model architecture for the GNN-based node-pair-level prediction for PCB component addition. The GNN output representations of the net nodes are passed to the pre-filter MLP, which identifies individual net nodes with a high probability of serving as a connection point to the new component. The filtered node representations are concatenated pairwise and passed to a second MLP, which performs a task-specific prediction on each of these node pairs. In this example, the node pair MLP performs a multiclass classification to predict the new component's position and type (R or C).} \label{fig:model}
\end{figure}\\\\
\textbf{Model Architecture.} Our overall model is depicted in Fig.~\ref{fig:model}. The input to the model is the graph representation of a PCB schematic, as described in the previous section. It contains the input node features, a combination of the node type and name, as well as edge attributes containing the pin name embeddings. An arbitrary GNN can be used to process the graph and calculate hidden node representations. The final node representations are passed to the pre-filtering MLP, which assigns a probability score to each net node representation. All net nodes with a probability score higher than a predefined threshold $\theta$ are passed to the node pair prediction module. This module is another MLP that performs a final prediction on the concatenation of two node representations. Thereby, this prediction is only performed for all possible pairs of nodes that were not sorted out by the pre-filtering MLP. Note that the threshold $\theta$ can be treated as an additional hyperparameter.\\\\
\textbf{Task-Specific Model Output.} The final output of the node pair prediction depends on the specific learning task. It could be a binary classification if the type of the new component is known, and a multiclass classification if there are multiple possible classes of components. Finally, it could also be a regression if the task is to predict the number of parallel components to insert between a specific pair of nodes.

\section{Experiments}\label{sec:experiments}
We validate our approach by performing experiments on three specific use cases, each corresponding to the addition of a different component for the optimization of the PCB design. First, we perform a binary classification to predict the positions of additional pull-up- and pull-down resistors. Second, we train a model on inserting RC filters on digital reset pins, which corresponds to a multiclass classification since the model has to distinguish also between resistor and capacitor. Finally, we investigate the addition of decoupling capacitors, which we treat both as a simplified binary classification task ("Add at least one capacitor or not") and as a regression task ("How many parallel capacitors should be placed at this position?").

\subsection{Datasets, Models, and Experimental Setting}
\textbf{Datasets.} We generated three large datasets of labeled PCB schematics, one for each optimization task. The schematics were optimized and labeled manually by human experts in electrical engineering. Thereby, each pair of net nodes in each circuit of the dataset received a label $\hat{y}_{\text{pair}}$ according to the specific task (binary, multiclass, or regression). Additionally, we generated binary labels $\hat{y}_{\text{node}}$ for each net node indicating whether it serves as a connection point for at least one new component or not. These single-node labels are used to train the pre-filter MLP. Statistics of the three resulting datasets can be found in Tab.~\ref{tab:datasets}. All three datasets contain a large number of unique graphs of varying sizes ranging from 6 nodes up to huge graphs with more than 700 nodes. On average, between one and three new nodes are added per graph. This means that only a small fraction of all possible net node pairs receive a positive label, resulting in a strong class imbalance.
\begin{table}[t]
\caption{Dataset statistics. Percentages are calculated with respect to the average number of nodes per graph.}\label{tab:datasets}
\begin{center}
\begin{tabular}{l|c|c|c}
\toprule
Dataset                          & ~Pull-Ups/-Downs~ & ~~~~RC-Filters~~~~ & ~Decoupling Caps.~  \\
\midrule
No. of Graph Samples                   & 2396            & 849           & 944            \\
Avg. No. of Nodes               & 122.6           & 123.5         & 100.1          \\
Min. No. of Nodes               & 7               & 7             & 6              \\
Max. No. of Nodes               & 702             & 637           & 500            \\
Avg. No. of Edges               & 174.4           & 175.7         & 139.0          \\
Avg. No. of Added Nodes         & 2.8   (2.3\%)          & 2.1  (1.7\%)         & 1.4 (1.4\%)           \\
\bottomrule
\end{tabular}
\end{center}
\end{table}\\\\
\textbf{GNN Models.} In all experiments, we apply our node pair prediction model with different GNN backbones. Thereby, we consider the baseline models GCN~\citep{kipf2017semi} and GIN~\citep{xu2019powerful} as well as GINe~\citep{Hu2020Strategies}, a modification of GIN that includes edge attributes. Furthermore, we experiment with several widely used attention-based GNN models: GAT~\citep{velivckovic2018graph}, GATv2~\citep{brody2021attentive}, and GraphTransformer (GT)~\citep{shi2020masked}. All three models are taking edge attributes into account. As an additional baseline, we also compare to an "MLP-only" version of our model, where the pre-filter MLP and the node pair MLP are applied directly to the input node features without any message-passing layers in between. In this way, we can specifically investigate the influence of the graph structure on the model performance.\\\\
\textbf{Experimental Setting.} We split each dataset into train, validation, and test sets with ratios 80/10/10\% and perform a 9-fold cross-validation. Additionally, we perform a small grid search for hyperparameter optimization (see Tab.~\ref{tab:hyperparameters} for details). For each training run, we use the AdamW optimizer \citep{loshchilov2018decoupled}, a batch size of 128, and perform early stopping with a patience of 20 epochs using the task-specific evaluation metric. The loss function is composed of a binary cross-entropy loss term $BCE$ responsible for the training of the pre-filter MLP, and a task-specific node pair term $L_{\text{task}}$ for the training of the node pair MLP:
\begin{align}
    L = BCE(y_{\text{node}}, \hat{y}_{\text{node}}) + L_{\text{Task}}(y_{\text{pair}}, \hat{y}_{\text{pair}}).
\end{align}
Note that both loss terms also contribute to training the GNN layers.
\begin{table}[t]
\caption{Overview of the hyperparameters used in the experiments.}\label{tab:hyperparameters}
\begin{center}
\begin{tabular}{lr}
\toprule
Hyperparameter     & Values                  \\
\midrule
Hid. Dimension     & 16, 32, 64              \\
Num. of GNN Layers & 1, 2, 3                 \\
Attention Heads    & 1, 4                    \\
Learning Rate      & 0.001, 0.0005, 0.0001   \\
Threshold $\theta$ & 0.0, 0.1, ..., 0.7 \\
\bottomrule
\end{tabular}
\end{center}
\end{table}

\subsection{Optimization Task 1: Adding Pull-Up and Pull-Down Resistors}

\textbf{Technical Background.} Pull-up and pull-down resistors are two-terminal components placed between one specific net on the design and, in the case of pull-ups, a supply net, and in the case of pull-downs, a ground net. When placed, these components help to ensure that the voltage on the selected design net is brought to a defined known value in case the design net is left floating. There are many possible reasons for the usage of pull-ups and pull-downs, including functionality, reliability, and best practices. In a functional safety example, if a MOSFET transistor is not being actively driven, a pull-down resistor can ensure that the transistor is definitely turned off.\\\\
\textbf{Task-Specific Model.} We approach this optimization task as a binary classification on the node pair level, where a positive label corresponds to the placement of a pull-up or pull-down resistor. We do not differentiate explicitly between pull-up and pull-down resistors, since the component is the same in both cases and the function of the resistor follows straight from its position within the circuit. Thus, the node pair MLP has a single output node and for the loss term $L_{\text{Task}}$, we use another binary cross-entropy loss function:
\begin{align}
    L_{\text{Task}}(y_{\text{pair}}, \hat{y}_{\text{pair}}) = BCE(y_{\text{pair}}, \hat{y}_{\text{pair}}).
\end{align}
Due to the extreme class imbalance in the dataset, we use the Area Under the Precision-Recall Curve (AUPRC) as our evaluation metric, which is a robust metric for scenarios with an underrepresented positive class that focuses on the trade-off between precision and recall. We calculate the AUPRC over the predictions for all possible pairs of net nodes in each graph. Thereby, all predictions for node pairs that are not evaluated by the node pair MLP (because at least one of the two nodes was sorted out by the pre-filter MLP) are set to negative labels (no component insertion at this position). Therefore, the AUPRC metric reflects both errors resulting from node pair misclassifications as well as errors resulting from incorrect filtering by the pre-filter MLP.
\begin{figure}[t]
\centering
\includegraphics[width=0.9\textwidth]{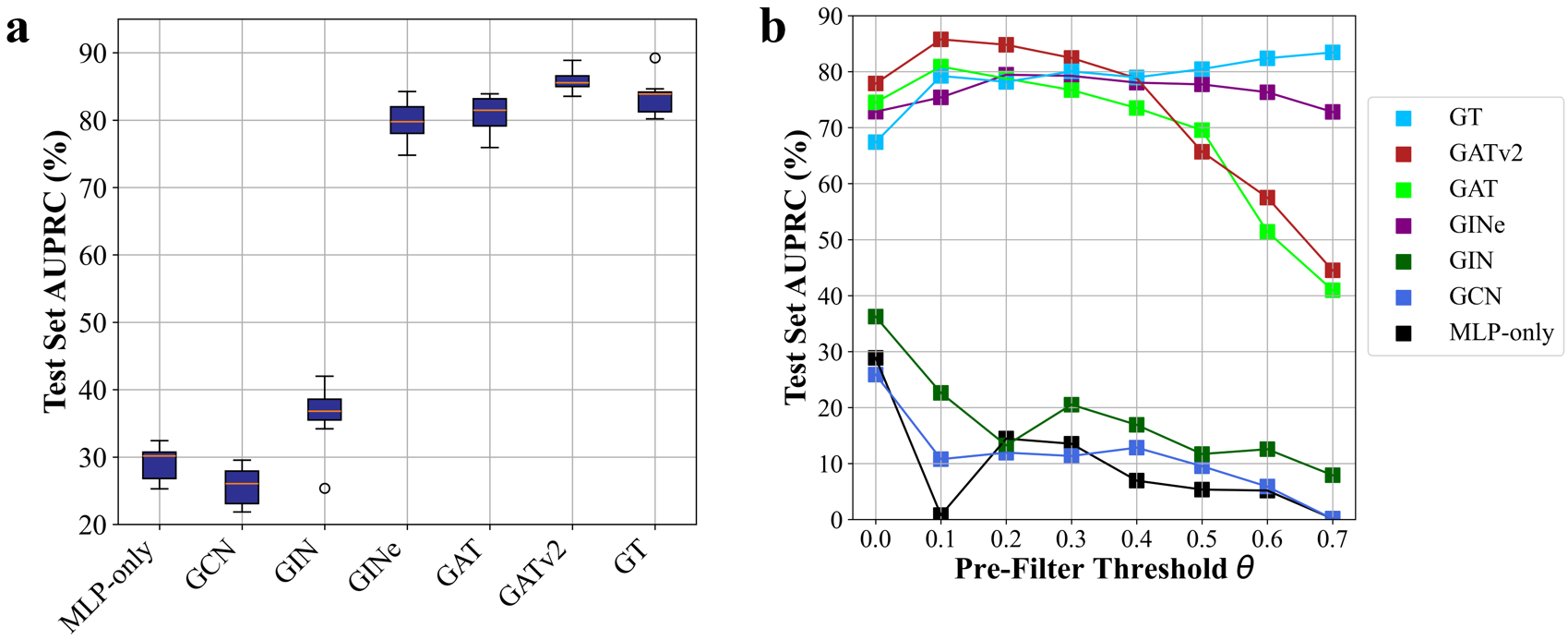}
\caption{\textbf{a} Area under the precision-recall curve (AUPRC) on the test set for the best hyperparameter configuration of all model variants for the pull-ups/-downs insertion task. \textbf{b} Sensitivity of the pull-ups/-downs insertion performance on the pre-filter threshold $\theta$ for all model variants.} \label{fig:pull_results}
\end{figure}\\\\
\textbf{Results.} Fig.~\hyperref[fig:pull_results]{\ref*{fig:pull_results}a} shows the AUPRC on the test set for all considered models. Thereby, we only consider the best hyperparameter configurations, which are reported in App.~\ref{ref:app_a}. First, it is noticeable that the model variants MLP-only, GCN and GIN show a much lower performance compared to all other models. These three models do not consider edge features, which appear to be very important for this specific task. A possible reason for this is that the names of the target connection points are sometimes chosen generically (e.g., "N\$2"), whereas the names of the connected pins, i.e., the edge attributes, may reveal more information on their function. The more advanced GNN variants GINe, GAT, GATv2, and GT all consider edge attributes and exhibit a much higher AUPRC of more than 80~$\%$. Among these models, GATv2 performs best with an AUPRC of $85.8~\%$.\\
The influence of the pre-filter threshold $\theta$ on the task performance is investigated in Fig.~\hyperref[fig:pull_results]{\ref*{fig:pull_results}b}. For the models with an edge-feature dependency, the performance drops when no pre-filter is used ($\theta = 0$). This underlines the importance of the pre-filter beyond computational efficiency: It also stabilizes the training process and increases the performance. The best performances are mostly achieved with a small threshold between 0.1 and 0.3, whereas the AUPRC decreases steadily for higher $\theta$.\\
These results indicate that it is better to not sort out too many nodes in the pre-filter step, since the node pair MLP is capable of correcting false positive predictions by the pre-filter. However, if the pre-filter MLP has too many false negatives (high $\theta$), the associated node pairs are not evaluated by the node pair MLP anymore and are instead automatically set to a negative label. Only the GT model appears to be an exception here: The test set performance increases for higher $\theta$ due to very accurate pre-filtering.\\
The models without edge-feature dependency (MLP-only, GCN, and GIN) often do not converge and show very low performance. In this case, the AUPRC decreases even more when a pre-filter is used, because these models do not learn sufficiently meaningful node representations. Thus, the pre-filter MLP predicts too many false negatives that the node pair MLP cannot correct anymore.

\subsection{Optimization Task 2: Adding RC Filters on Digital Reset Pins}

\textbf{Technical Background.} Circuit schematics containing digital logic, especially programmable devices such as microprocessors and microcontrollers, require the special handling of the reset signal pin to prevent unintentional reset events, leading to functional issues in the final product. Commonly, reset pins are built to be "active-low", meaning they should be tied to the supply voltage to stay inactive. However, due to noise created by power supplies and the processor, voltage glitches can be present in the supply line, creating these unintentional reset events. Therefore, it is advisable to always have a simple RC filter between the supply net and the reset pin, in order to smooth out the voltage level noise and prevent issues. An RC filter is composed of a resistor connected between the supply net and the reset net, and a capacitor connected between the reset net and the ground net.\\\\
\textbf{Task-Specific Model.} We treat the placement of the RC filters as a multiclass classification task, where the node pair MLP has three output nodes, each corresponding to one of the three possible class labels: Resistor, capacitor, or none. For the loss term $L_{\text{task}}$, we use a cross-entropy loss function:
\begin{align}
    L_{\text{Task}}(y_{\text{pair}}, \hat{y}_{\text{pair}}) = CEL(y_{\text{pair}}, \hat{y}_{\text{pair}}).
\end{align}
As an evaluation metric, we utilize the AUPRC with macro-averaging, i.e., the unweighted mean of the separate metrics for all three classes.
\begin{figure}[!t]
\centering
\includegraphics[width=0.9\textwidth]{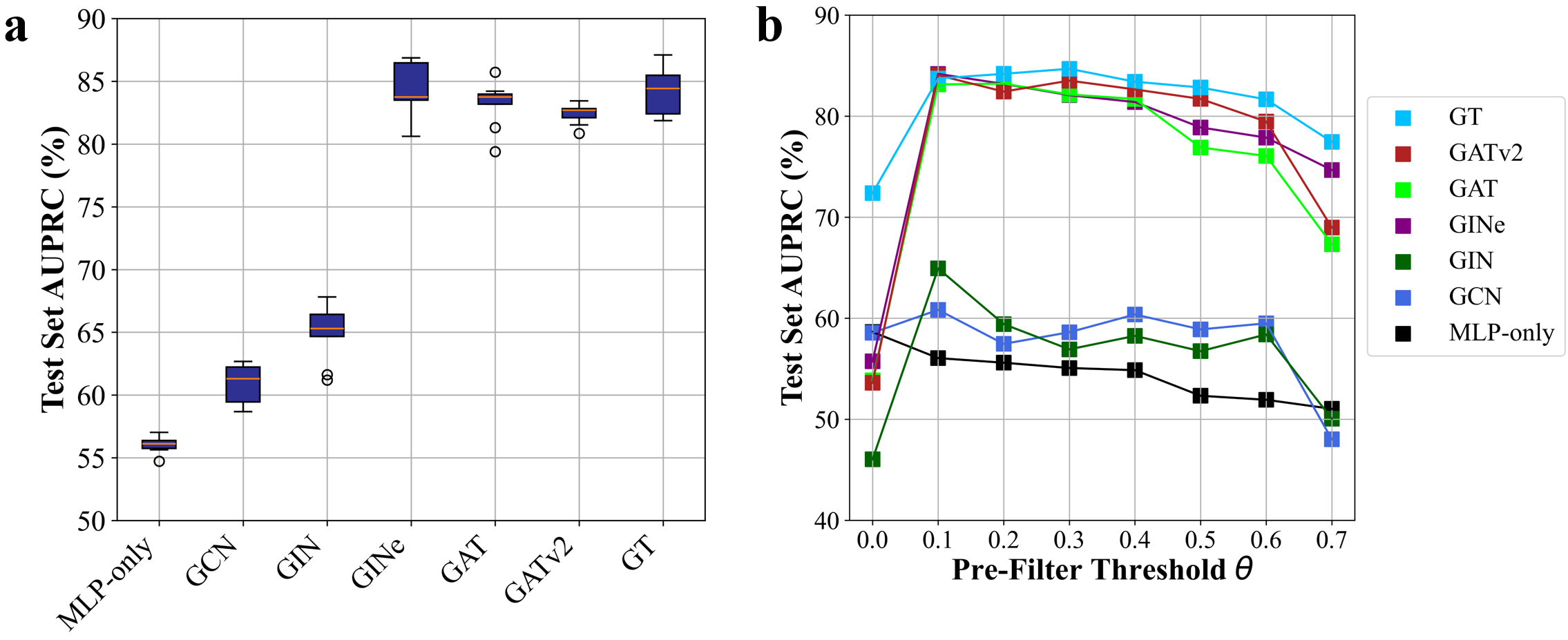}
\caption{\textbf{a} Area under the precision-recall curve (AUPRC) with macro-averaging on the test set for the best hyperparameter configuration of all model variants for the RC filter insertion task. \textbf{b} Sensitivity of the RC filter insertion performance on the pre-filter threshold $\theta$ for all model variants.} \label{fig:reset_results}
\end{figure}\\\\
\textbf{Results.} Fig.~\hyperref[fig:reset_results]{\ref*{fig:reset_results}a} shows the AUPRC on the test set for all considered models. First, it is noticeable that the model variants MLP-only, GCN and GIN again show a much lower performance compared to all other models, emphasizing the importance of the edge attributes for the identification of reset pins. Overall, the MLP-only variant shows the lowest performance of all models, which underlines the importance of the graph structure for this learning task. The attention-based GNN variants GAT, GATv2, and GT, as well as the GINe model, exhibit a very similar performance of nearly 85~$\%$ AUPRC.\\
The influence of the pre-filter threshold $\theta$ on the task performance is investigated in Fig.~\hyperref[fig:reset_results]{\ref*{fig:reset_results}b}. For all models, except the MLP-only variant, the performance drops significantly when no pre-filter is used ($\theta = 0$). Furthermore, the models with edge-feature dependency exhibit a slow performance decrease for increasing $\theta$.

\subsection{Optimization Task 3: Adding Decoupling Capacitors to Supply}

\textbf{Technical Background.} Most electronic circuits have at least one supply net, used to provide power to integrated circuits, and one ground net, which is the current return net back to the supply. Both supply and ground nets are shared among many components in an electronic design, and therefore it is important to make sure the operation of one integrated circuit does not interfere with the operation of another one. To prevent these undesired interactions through the supply net, decoupling capacitors are placed. They are connected as close as possible between supply and ground nets, which can "short-circuit" high-frequency noise created by digital integrated circuits and make sure both supply and ground nets remain clean. This helps to contain this noise and avoid interferences, as well as to reduce the levels of electromagnetic emission from the final device.\\\\
\textbf{Task-Specific Model.} We approach this optimization task from two perspectives simultaneously. First, we consider it as a binary classification task with the goal of predicting whether at least one decoupling capacitor has to be inserted between a pair of net nodes or not. For different reasons, however, engineers are often placing multiple decoupling capacitors in parallel, e.g., to ensure stability over a wider range of frequencies. Therefore, we also perform a regression and predict the exact number of parallel capacitors that engineers would insert at a certain position in the circuit.\\
We utilize two separate node pair MLPs for this task, one for the binary classification with outputs $z_{\text{pair}}$ and one for the regression with outputs $y_{\text{pair}}$. For the binary classification, we use the binary cross-entropy loss function and for the regression, we utilize the mean-squared error. We further assign a weight $\alpha$ to the regression loss term to control its influence on the training process:
\begin{align}
     L_{\text{Task}}(z_{\text{pair}}, y_{\text{pair}}, \hat{y}_{\text{pair}}) = BCE(z_{\text{pair}}, \hat{z}_{\text{pair}}) + \alpha \cdot MSE(y_{\text{pair}}, \hat{y}_{\text{pair}}).
\end{align}
Thereby, the labels $\hat{z}_{\text{pair}}$ are computed as
\begin{align}
    \hat{z}_{\text{pair}} = \begin{cases} 
    1, ~~~\hat{y}_{\text{pair}} \geq 1 \\
    0, ~~~\text{otherwise.}
    \end{cases}
\end{align}
\begin{figure}[t]
\centering
\includegraphics[width=0.9\textwidth]{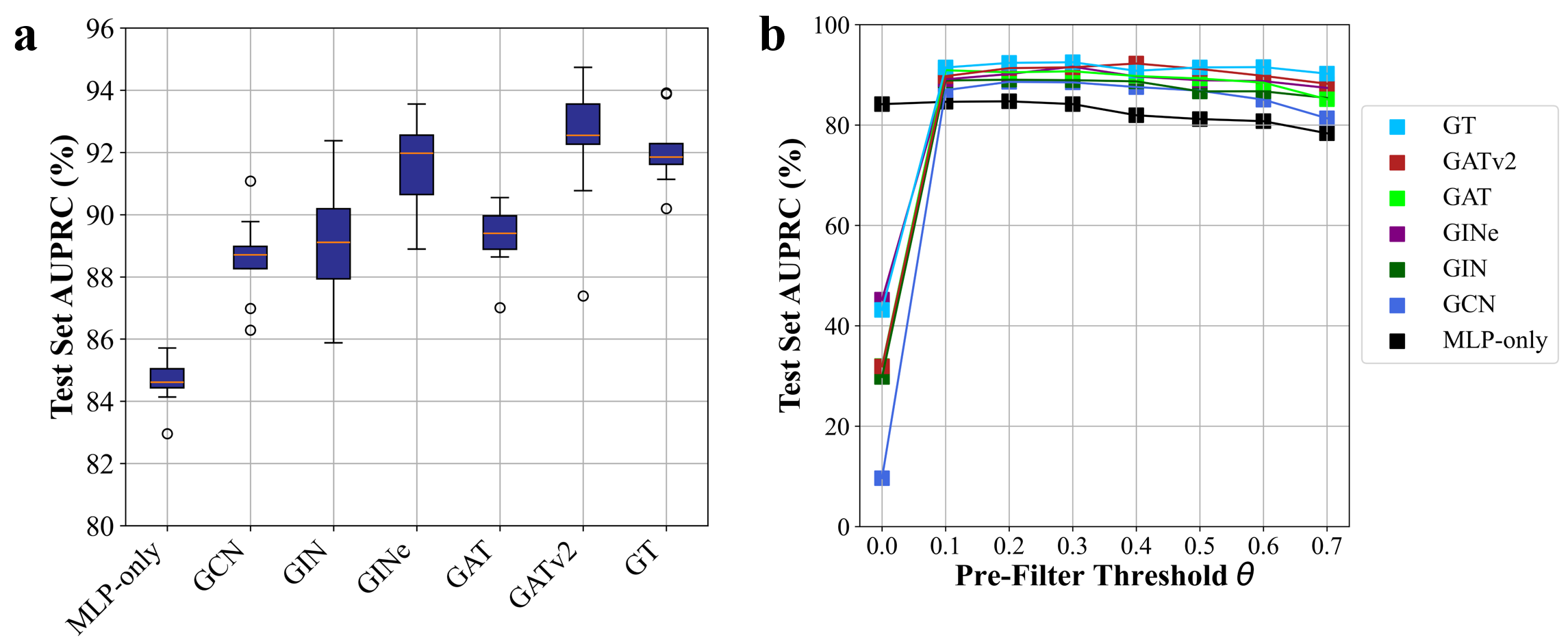}
\caption{\textbf{a} Area under the precision-recall curve (AUPRC) on the test set for the best hyperparameter configuration of all model variants for the decoupling capacitor insertion task (treated as a binary classification). \textbf{b} Sensitivity of the decoupling capacitor insertion performance on the pre-filter threshold $\theta$ for all model variants.} \label{fig:decoupling_results}
\end{figure}\\\\
\textbf{Binary Classification Results.} For the evaluation of the binary classification, we utilize the AUPRC metric again. Fig.~\hyperref[fig:decoupling_results]{\ref*{fig:decoupling_results}a} shows the AUPRC on the test set for all considered models. Again, the MLP-only variant shows the lowest performance among all models, although it performs much better than on the other tasks. This indicates that the graph structure and local node neighborhood are less relevant for the decoupling capacitor insertion task. Furthermore, GNNs that include edge attributes (GINe, GAT, GATv2, and GT) only show a slightly better performance compared to GCN and GIN. Therefore, we conclude that the node features, i.e., node types and names, are the most important features for the prediction of decoupling capacitor additions. A possible reason for this is that supply and ground nets, which are the relevant connection points for decoupling capacitors, mostly have well-defined names, making them easily identifiable by their node features alone.\\
Fig.~\hyperref[fig:decoupling_results]{\ref*{fig:decoupling_results}b} shows the sensitivity of the test set AUPRC on the pre-filter threshold $\theta$. Here, the performance drops drastically when no pre-filter is used ($\theta = 0$) and decreases slightly for increasing $\theta$. Only the MLP-only variant shows an acceptable performance without any pre-filter.
\begin{figure}[!ht]
\centering
\includegraphics[width=0.45\linewidth]{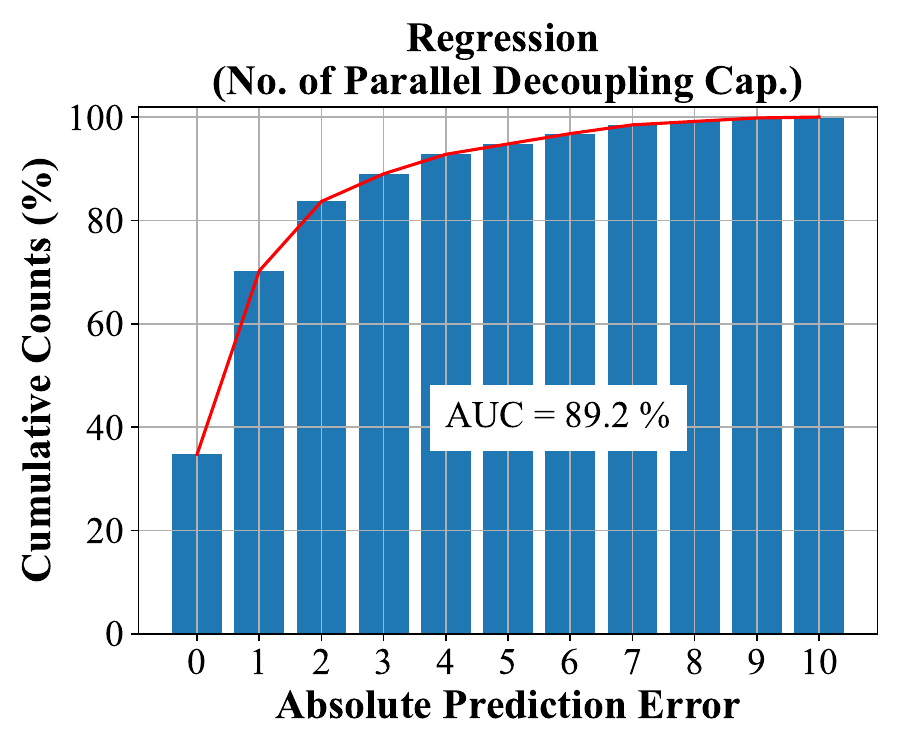}
\caption{Cumulative counts of absolute errors of the GATv2 model variant for the prediction of the number of parallel decoupling capacitors to insert at a given position in the circuit. In approx. 70$\%$ of the cases, the prediction deviates from the ground truth by at most 1.} \label{fig:decoupling_reg}
\end{figure}\\\\
\textbf{Regression Results.} For the evaluation of the regression, we first round the outputs of the regression MLP to integers. Next, we calculate the absolute prediction error for all non-zero labels. Fig.~\ref{fig:decoupling_reg} shows the cumulative counts of absolute prediction errors for the GATv2 model variant with $\alpha = 0.1$. In approximately $70~\%$ of the cases, the predicted number of parallel decoupling capacitors at a given position deviates from the ground truth by at most 1. The overall area under the curve is $89.2~\%$, indicating that the model can give a good estimation of the number of parallel decoupling capacitors, despite lacking some information that engineers would take into account when solving this problem, e.g., component values or market availability of certain capacitors.

\section{Conclusion}

In this paper, we presented a Graph Neural Network (GNN)-based approach for automating the addition of new components in Printed Circuit Board (PCB) schematics to optimize their robustness and reliability. Thereby, we represented PCB schematics as bipartite graphs and trained a node-pair-level classification on three real-world PCB datasets manually labeled by human experts, each focusing on the addition of different optimizing components with high practical relevance: Pull-up/-down resistors, RC filters, and decoupling capacitors.\\
Our results show that GNNs, especially architectures that consider edge attributes, can learn meaningful representations of PCB schematics and provide a significant performance gain on all optimization tasks compared to the usage of pure multi-layer perceptrons. Furthermore, we found that pre-filtering promising connection points for the new components using a separate multi-layer perceptron stabilizes the training process and increases the overall model performance. In summary, the accuracy of our approach is high enough to increase the automation level of the PCB design optimization process and support engineers in developing more durable electronic devices more efficiently, leading to considerable time and cost savings.\\
In the future, we want to further enhance the model performance by investigating the usage of other graph types for the representation of PCB schematics (e.g., hypergraphs) as well as more specialized GNN architectures such as heterogeneous GNNs. Furthermore, we want to consider additional information from the PCB schematics that has not been exploited so far, e.g., component values like resistances or capacities. Finally, we plan to extend the scope of our graph-based approach to other PCB design optimization tasks involving not only the addition of new components but also the merging, splitting, or removal of existing ones.

\section{Acknowledgments} This study was funded by the  German Federal Ministry of Research, Technology, and Space (funding code 16ME0877). We also thank Clara Holzhüter for her helpful feedback.

\newpage
\appendix
\section{Results of the Hyperparameter Optimizations}
\label{ref:app_a}

Here, we report the results of the hyperparameter tuning, i.e., the hyperparameter configuration for each model and optimization task that yielded the lowest validation set error.

\begin{table}[!ht]
\caption{Best hyperparameter configurations of all model variants for the pull-up/pull-down optimization task.}
\begin{center}
\begin{tabular}{l|c|c|c|c|c|c|c}
\toprule
Model              & MLP-only & ~~~GCN~~~   & ~~~GIN~~~   & ~~GINe~~  & ~~~GAT~~~   & ~~GATv2~~ & ~~~~GT~~~~    \\
\midrule
No. of GNN Layers  & 0        & 3     & 3     & 3     & 3     & 3     & 3     \\
Hidden Dimension   & 64       & 64    & 64    & 64    & 64    & 64    & 64    \\
Attention Heads    & -        & -     & -     & -     & 4     & 4     & 4     \\
Learning Rate      & 0.001    & 0.001 & 0.001 & 0.001 & 0.001 & 0.001 & 0.001 \\
Threshold $\theta$ & 0.0      & 0.0   & 0.0   & 0.2   & 0.1   & 0.1   & 0.7  \\
\bottomrule
\end{tabular}
\end{center}
\end{table}

\begin{table}[!ht]
\caption{Best hyperparameter configurations of all model variants for the RC filter optimization task.}
\begin{center}
\begin{tabular}{l|c|c|c|c|c|c|c}
\toprule
Model              & MLP-only & ~~~GCN~~~   & ~~~GIN~~~   & ~~GINe~~  & ~~~GAT~~~   & ~~GATv2~~ & ~~~~GT~~~~    \\
\midrule
No. of GNN Layers  & 0        & 3     & 3     & 2     & 1     & 1     & 3     \\
Hidden Dimension   & 64       & 64    & 64    & 64    & 64    & 64    & 64    \\
Attention Heads    & -        & -     & -     & -     & 4     & 4     & 1     \\
Learning Rate      & 0.001    & 0.001 & 0.0005 & 0.001 & 0.001 & 0.001 & 0.001 \\
Threshold $\theta$ & 0.0      & 0.1   & 0.1   & 0.1   & 0.2   & 0.1   & 0.1  \\
\bottomrule
\end{tabular}
\end{center}
\end{table}

\begin{table}[!ht]
\caption{Best hyperparameter configurations of all model variants for the decoupling capacitor optimization task.}
\begin{center}
\begin{tabular}{l|c|c|c|c|c|c|c}
\toprule
Model              & MLP-only & ~~~GCN~~~   & ~~~GIN~~~   & ~~GINe~~  & ~~~GAT~~~   & ~~GATv2~~ & ~~~~GT~~~~    \\
\midrule
No. of GNN Layers  & 0        & 2     & 2     & 1     & 1     & 1     & 3     \\
Hidden Dimension   & 64       & 64    & 64    & 64    & 64    & 64    & 64    \\
Attention Heads    & -        & -     & -     & -     & 4     & 4     & 4     \\
Learning Rate      & 0.001    & 0.001 & 0.001 & 0.001 & 0.001 & 0.001 & 0.001 \\
Threshold $\theta$ & 0.1      & 0.2   & 0.1   & 0.3   & 0.5   & 0.4   & 0.1  \\
\bottomrule
\end{tabular}
\end{center}
\end{table}

\section{A Note on the Symmetry of the Node Pair Predictions}

The input to the node pair MLP is a \emph{node pair representation}, i.e., the concatenation of two individual node representations. Since we represent pins as undirected edges, there is no preferred order for this concatenation and in general, the two possible orders may result in different MLP outputs. However, during training, we always consider both possible concatenation orders for the node pair representations. In this way, the model learns that both orders are equivalent since they always have the same label. We can verify the symmetry of the node pair predictions by calculating the correlation between the upper and lower triangles of the output matrices for the trained models. In all our experiments, this correlation is above 99.9$~\%$, indicating that our models are nearly invariant with respect to permutation of the nodes in the node pair representation.

\end{document}